\documentclass[10pt,twocolumn,letterpaper]{article}

\usepackage[pagenumbers]{cvpr} 
\usepackage{pbalance}

\definecolor{cvprblue}{rgb}{0.21,0.49,0.74}
\usepackage[pagebackref,breaklinks,colorlinks,allcolors=cvprblue]{hyperref}
\usepackage[accsupp]{axessibility}  

\def\paperID{31}
\def\confName{CVPR}
\def\confYear{2026}

\title{Frequency-Guided Fusion For RGB-Thermal Semantic Segmentation}

\author{İsmail Emre Canıtez, Özgür Erkent\\
Hacettepe University\\
{\tt\small \{emrecanitez, ozgurerkent\}@hacettepe.edu.tr}
}
\begin{document}
\maketitle
\begin{abstract}
Semantic segmentation in complex environments such as urban driving 
scenes remains challenging under adverse lighting conditions, where RGB 
images alone provide insufficient information. RGB-Thermal fusion 
leverages the complementary strengths of visible and infrared imagery 
to improve scene understanding; however, effectively integrating these 
heterogeneous modalities at varying levels of feature abstraction 
remains an open problem. 
In this paper, we propose a multi-modal fusion architecture built upon 
dual ConvNeXt V2 backbones that employs stage-wise, modality-adaptive 
fusion strategies. For early-stage features, we introduce a 
Frequency-Based Fusion Module that decomposes infrared features into 
low- and high-frequency components via Gaussian filtering, applies 
dual-branch spatial attention to selectively emphasize thermal patterns 
and fine-grained boundaries, and integrates them with RGB features 
through a confidence-gated residual mechanism. For late-stage features, 
we design a semantic fusion module with cross-modal attention and 
multi-scale depthwise convolutions to capture semantic correspondences 
across modalities. The fused features are decoded via a PANet-style 
bidirectional decoder with deep supervision. Experiments on MFNet and 
PST900 demonstrate that our lightest variant achieves 61.73\% and 
86.24\% mIoU, respectively, with only 35.43M parameters, outperforming 
recent methods while using substantially fewer parameters and lower 
computational cost. Code is available at \url{https://github.com/ismailemrecntz/VISIBLE-INFRARED-SENSOR-FUSION}.
\end{abstract}
    
\section{Introduction}
\label{sec:intro}
Semantic segmentation, which assigns a class label to every pixel in an image, is a fundamental task in computer vision with direct implications for safety-critical applications such as autonomous driving \cite{chen2018deeplab, zhao2017pspnet} and robotic navigation \cite{milioto2019rangenet}. Although deep learning methods operating on RGB imagery have achieved substantial progress \cite{he2016deep, ronneberger2015unet}, their performance degrades significantly under adverse lighting conditions such as nighttime, heavy shadows, and overexposure. Thermal imaging provides a valuable complement: unlike visible-light cameras, thermal sensors capture emitted radiation that remains largely invariant to ambient illumination, rendering pedestrians and vehicles clearly distinguishable even in complete darkness. However, infrared images lack the fine texture, color, and structural detail present in well-lit RGB images. This complementarity motivates the development of RGB-Thermal (RGB-T) fusion methods that jointly leverage both modalities for robust scene understanding across diverse lighting conditions \cite{RTFNet, GMNet}.

\begin{figure}[!t]
  \centering
   \includegraphics[width=\linewidth]{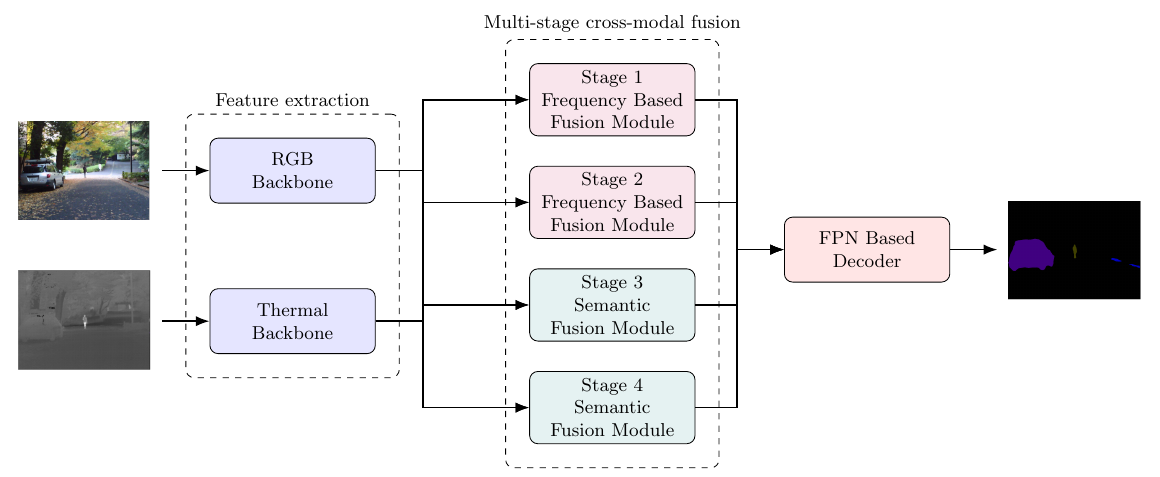}
    \caption{Overview of the proposed RGB–Thermal semantic segmentation 
    framework.}

    \label{fig:model}
\end{figure}

The two modalities possess distinct statistical properties—RGB features encode appearance and color, while thermal features primarily represent temperature distributions—and simple concatenation or summation often yields suboptimal results \cite{CMX}. Moreover, the relevance of each modality varies with scene conditions: RGB provides richer cues in well-lit environments, whereas the thermal modality becomes more informative at night. An effective fusion strategy must therefore adapt to the relative reliability of each modality \cite{EGFNet}. Existing middle-fusion approaches, while most promising, typically apply the same fusion operation uniformly across all feature levels, neglecting that low-level features capture fine-grained spatial details while high-level features encode semantic context. Several recent methods achieve competitive accuracy but incur substantial computational overhead \cite{Sigma, CMX}, limiting their suitability for resource-constrained deployment.
A separate line of research has explored the frequency properties of visual features for multi-modal fusion. However, frequency-aware processing in the context of RGB-T segmentation remains notably underexplored. The few existing works—SGFNet \cite{zhang2025spectral}, which applies DCT-based high-frequency extraction, and Wavelet-CNet \cite{waveletcnet}, which uses wavelet cross fusion—operate in the transform domain and process both modalities symmetrically. None of these explicitly leverages the unique physical properties of infrared imagery: thermal images encode surface temperature distributions that naturally lend themselves to frequency decomposition, where low-frequency content corresponds to broad thermal regions and high-frequency content captures thermally distinct boundaries. We argue that selectively decomposing \emph{infrared} features and using the resulting components to complement RGB representations is both physically motivated and architecturally effective.

Building on these observations, we present a hierarchical dual-encoder architecture that adapts its fusion strategy to the characteristics of each feature level. Our architecture utilizes ConvNeXt V2 \cite{convnextv2} as the backbone encoder for both modalities—to our knowledge, the first application of ConvNeXt V2 in RGB-T segmentation. Its fully convolutional masked autoencoder (FCMAE) pre-training yields robust hierarchical representations that are particularly well-suited for multi-modal fusion, as the learned features are inherently resilient to partial input corruption—a property analogous to the degraded RGB signal encountered under adverse lighting. For early-stage features (stages 1–2), we introduce a \textbf{Frequency-Based Fusion Module} that decomposes infrared features into low- and high-frequency components, applies dual-branch spatial attention with large-kernel (7$\times$7) attention for thermal patterns and small-kernel (3$\times$3) attention for fine edges, and integrates them with RGB features through a confidence-gated residual mechanism that adapts to RGB reliability. For late-stage features (stages 3–4), we employ a \textbf{Semantic Fusion Module}  combining cross-modal channel attention, multi-scale depthwise convolutions, and joint channel-spatial attention. The fused multi-scale features are decoded via a PANet-style \cite{PANet} bidirectional decoder with deep supervision for improved gradient flow.
Our main contributions are as follows:
\begin{itemize}
\item We propose a stage-wise fusion strategy that explicitly differentiates 
      the fusion mechanism according to the nature of features at each 
      backbone stage, applying frequency-aware fusion for low-level features 
      and semantic cross-gated fusion for high-level features.
\item We introduce a Frequency-Based Fusion Module that, unlike prior 
      spectral approaches operating symmetrically on both modalities, 
      performs spatial-domain frequency decomposition exclusively on infrared 
      features at early encoding stages and uses the decomposed components 
      to selectively enhance RGB representations through a confidence-gated 
      residual mechanism—a perspective not explored in prior RGB-T 
      segmentation methods.
\item We adopt a PANet-style bidirectional decoder with deep supervision 
      that improves multi-scale prediction and supports better gradient 
      flow during training.
\item Extensive experiments on MFNet and PST900 demonstrate that our 
      approach achieves state-of-the-art accuracy on MFNet with substantially 
      fewer parameters and lower computational cost compared to recent 
      methods. Comprehensive ablation studies confirm the individual 
      contribution of thermal-specific frequency fusion, stage-adaptive 
      design, and deep supervision.

\end{itemize}




\section{Related Work}
\label{sec:relatedwork}

\subsection{RGB-Thermal Semantic Segmentation}

RGB-Thermal fusion for semantic segmentation has been extensively studied, with methods typically categorized by where and how the modalities interact. Early fusion approaches concatenate or add RGB and thermal inputs before a shared encoder, limiting the ability to learn modality-specific representations. FuseSeg~\cite{FuseSeg} mitigates this with dual DenseNet encoders and a two-stage fusion strategy, though its hierarchical fusion remains relatively simple. Late fusion methods process each modality through separate pipelines and merge predictions only at the output level, sacrificing intermediate cross-modal interaction.

The most widely adopted paradigm employs dual encoders with middle fusion. MFNet~\cite{mfnet} introduced a compact two-stream architecture using mini-inception blocks with short-cut connections for urban scene understanding. RTFNet~\cite{RTFNet} proposed a deeper ResNet-based dual-encoder design where thermal features are injected into the RGB stream via element-wise addition at each stage. While effective, this uniform additive fusion does not account for the differing nature of information across feature hierarchies. Subsequent works introduced more sophisticated fusion mechanisms: ABMDRNet~\cite{ABMDRNet} employs a bridging-then-fusing strategy to first reduce modality differences before combining features; GMNet~\cite{GMNet} applies graded-feature multilabel learning for hierarchical fusion; and EGFNet~\cite{EGFNet} incorporates edge-aware guidance to improve boundary segmentation.

More recently, transformer-based architectures have been applied to RGB-T segmentation. CMX~\cite{CMX} introduces a cross-modal feature rectification module and a feature fusion module with cross-attention for long-range context exchange, achieving strong results across multiple RGB-X tasks. CMNeXt~\cite{CMNext} extends this paradigm to arbitrary-modal segmentation. Sigma~\cite{Sigma} employs a Siamese Mamba architecture for efficient multi-modal fusion. While these methods achieve competitive accuracy, they typically incur substantial computational costs, limiting their suitability for resource-constrained deployment. Our approach returns to a purely convolutional design using ConvNeXt V2~\cite{convnextv2} backbones, achieving comparable accuracy with lower computational overhead.

\subsection{Frequency-Domain Feature Fusion}

Frequency-domain analysis has been widely explored in single-modality computer vision tasks through transforms such as Fourier, DCT, and wavelet decompositions. However, its application to multi-modal RGB-T fusion remains in early stages. SGFNet~\cite{zhang2025spectral} is among the first works to explicitly model spectral characteristics for RGB-T segmentation. It employs DCT-based processing within a Spectral-aware Global Fusion module that operates on features from both modalities symmetrically, using ResNet-152 as its backbone. While SGFNet demonstrates the potential of frequency-aware processing, its approach operates in the transform domain and does not differentiate between the frequency characteristics of the two modalities. Wavelet-CNet~\cite{waveletcnet} explores wavelet-based cross fusion for RGB-T segmentation, similarly treating both modalities uniformly in the wavelet domain.

Our approach differs from these works in two fundamental ways. First, we perform frequency decomposition exclusively on the infrared modality in the spatial domain using Gaussian filtering and unsharp masking, motivated by the physical observation that thermal imagery encodes surface temperature distributions with naturally distinct low-frequency (broad thermal regions) and high-frequency (thermal boundaries) content. Second, we restrict frequency-aware fusion to the early encoding stages where structural details dominate, rather than applying it uniformly across all levels. This stage-specific design reflects the observation that frequency decomposition is most beneficial when spatial resolution is high and features retain fine-grained structural information, while deeper semantic features are better served by cross-modal gating mechanisms.

\subsection{Multi-Scale Decoding}

Feature Pyramid Networks (FPN)~\cite{lin2017feature} introduced a top-down pathway for constructing multi-scale feature maps, becoming a standard component in dense prediction tasks. PANet~\cite{PANet} extended this design with an additional bottom-up path, enabling information flow in both directions and improving the localization of small objects. Deep supervision~\cite{deepsup}, which applies auxiliary loss functions at intermediate network layers, has been shown to improve gradient flow and accelerate convergence, particularly for segmentation tasks with class imbalance. In our architecture, we adopt a PANet-style bidirectional decoder augmented with global context modulation and multi-level feature aggregation. Deep supervision at each pyramid level provides additional gradient signals during training.

\section{Proposed Method}
\label{sec3}

\begin{figure*}[!t]
    \centering
    \includegraphics[width=0.9\linewidth]{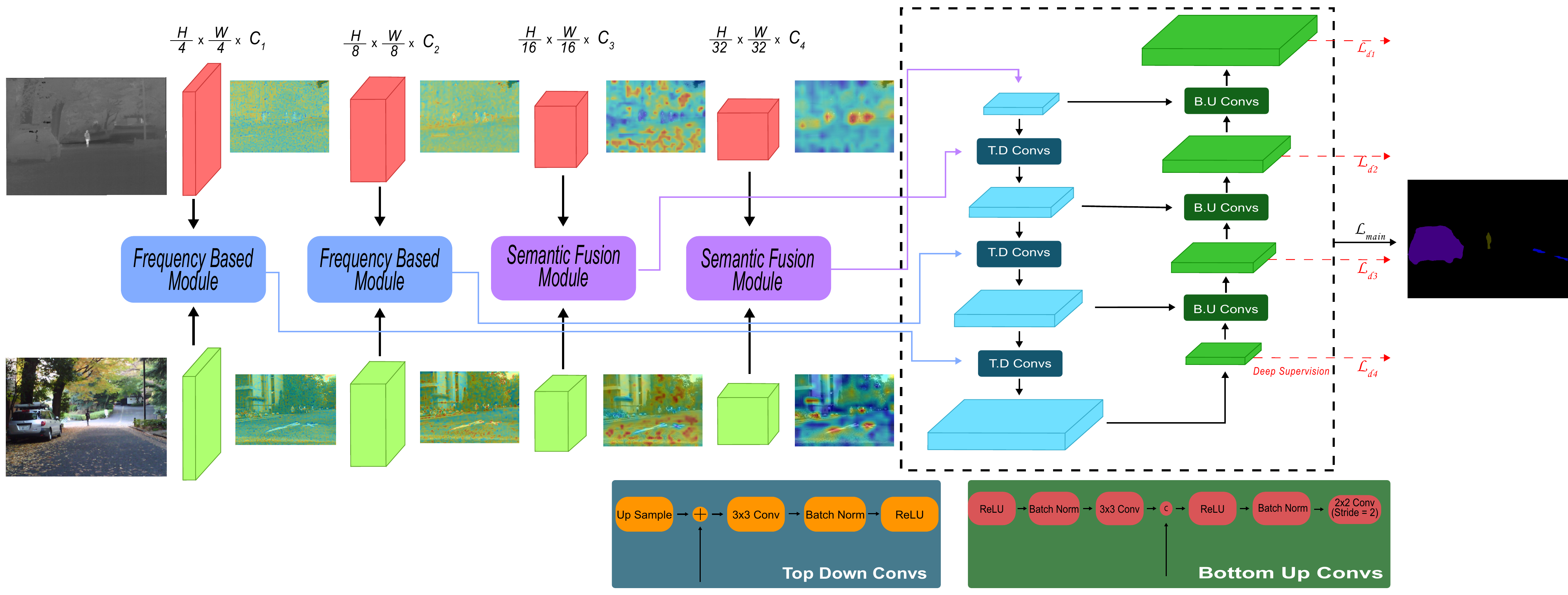}
    \caption{Overview of the proposed architecture. Dual ConvNeXt V2 encoders extract four-stage hierarchical features from RGB and thermal inputs. Stages 1--2 employ Frequency-Based Fusion; stages 3--4 use Semantic Fusion Module. A PANet-style bidirectional decoder with deep supervision produces the final segmentation.}
    \label{fig:architecture}
\end{figure*}

\subsection{Architecture Overview}
\label{sec:overview}

The overall architecture (\cref{fig:architecture}) follows a dual-stream encoder, stage-wise fusion, and bidirectional decoder paradigm. Given an RGB image $\mathbf{I}_{\text{rgb}} \in \mathbb{R}^{3 \times H \times W}$ and a thermal image $\mathbf{I}_{\text{ir}} \in \mathbb{R}^{1 \times H \times W}$, two parallel ConvNeXt~V2~\cite{convnextv2} backbones extract four levels of hierarchical features:
\begin{equation}
    \{R_i\}_{i=1}^{4} = \mathcal{E}_{\text{rgb}}(\mathbf{I}_{\text{rgb}}), \quad
    \{T_i\}_{i=1}^{4} = \mathcal{E}_{\text{ir}}(\mathbf{I}_{\text{ir}}),
\end{equation}
where $R_i, T_i \in \mathbb{R}^{C_i \times H_i \times W_i}$ with $C_i \in \{96, 192, 384, 768\}$ and spatial dimensions halved at each stage. Both branches use ConvNeXt~V2 pre-trained via the FCMAE protocol (Tiny variant by default; Nano and Base variants explored in \cref{sec:experiments}) on ImageNet-22k. The thermal encoder accepts a single-channel input by adapting the stem convolution while retaining pre-trained weights. Both encoders share the same architecture but maintain separate weights.

The fusion strategy is stage-dependent:
\begin{equation}
    F_i =
    \begin{cases}
        \mathcal{F}_{\text{freq}}(R_i, T_i), & i \in \{1, 2\}, \\
        \mathcal{F}_{\text{sem}}(R_i, T_i),  & i \in \{3, 4\}.
    \end{cases}
\end{equation}
The four fused maps $\{F_i\}_{i=1}^{4}$ are decoded by a PANet-style bidirectional decoder to produce the segmentation map $\hat{Y} \in \mathbb{R}^{K \times H \times W}$.

\begin{figure}[!t]
    \centering
    \includegraphics[width=0.70\linewidth]{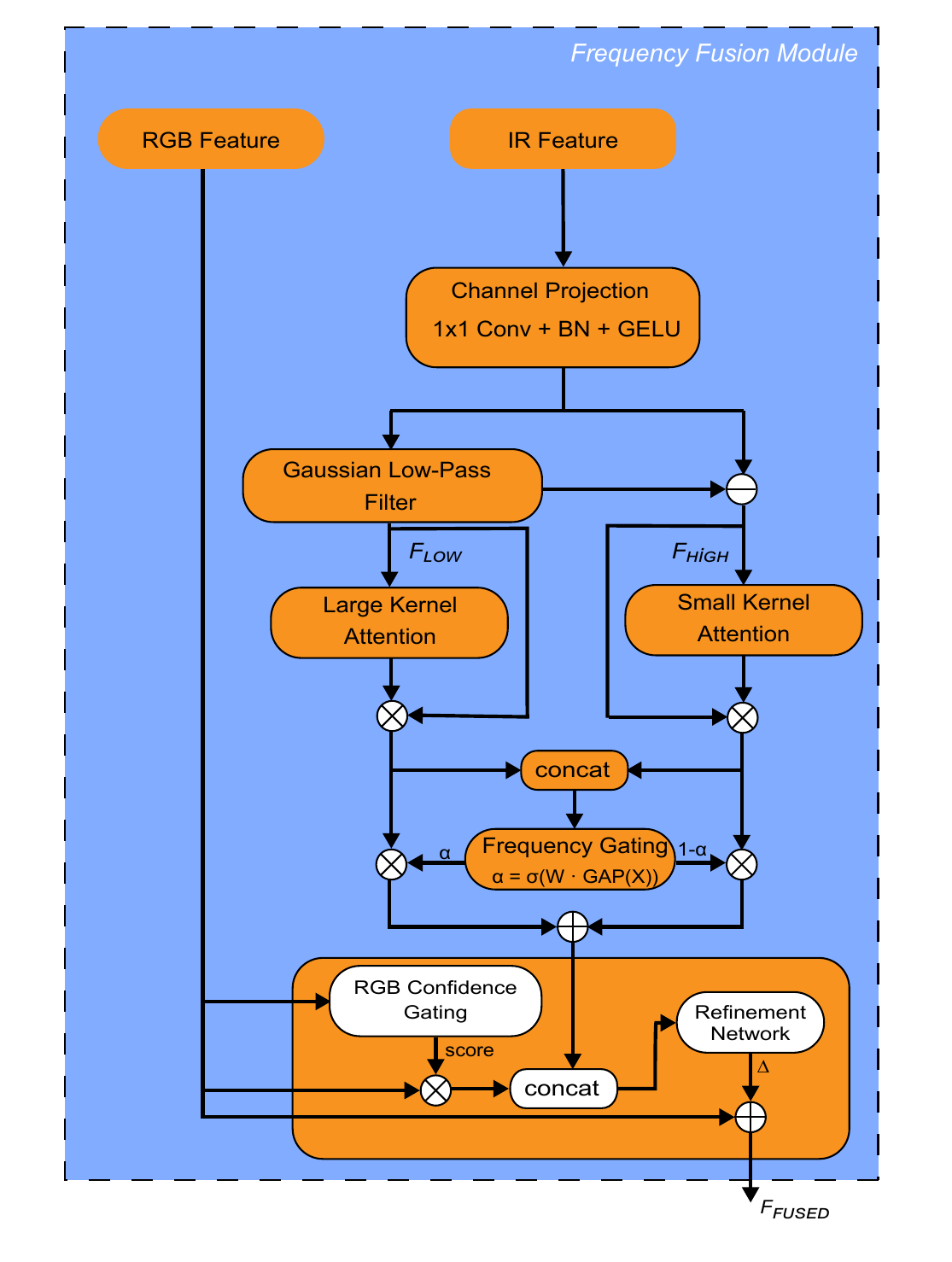}
    \caption{Frequency-Based Fusion Module (stages 1--2).}
    \label{fig:freq_fusion}
\end{figure}

\subsection{Frequency-Based Fusion Module}
\label{sec:freq_fusion}

At early stages, features encode fine-grained spatial details that benefit from explicit treatment of the infrared signal's frequency content (\cref{fig:freq_fusion}). The thermal features are first projected to the RGB channel space via $T_i' = \text{GELU}(\text{BN}(\text{Conv}_{1\times1}(T_i)))$.

\textbf{Frequency Decomposition.}
The projected features are decomposed using a fixed Gaussian kernel ($k{=}7$, $\sigma{=}2.0$) applied as a depthwise convolution:
\begin{equation}
    T_{\text{low}} = G_\sigma \circledast T_i', \quad T_{\text{high}} = T_i' - T_{\text{low}},
\end{equation}
separating broad thermal patterns (e.g., body heat) from thermally sharp boundaries (e.g., guardrail edges).

\textbf{Dual-Branch Spatial Attention.}
Each band is refined by a spatial attention mask $M(X) = \sigma(\text{Conv}_{k_s}[\text{AvgPool}_c(X) \| \text{MaxPool}_c(X)])$, using a large kernel ($k_s{=}7$) for low-frequency blobs and a small kernel ($k_s{=}3$) for high-frequency edges:
\begin{equation}
    \tilde{T}_{\text{low}} = T_{\text{low}} \odot M_7(T_{\text{low}}), \quad
    \tilde{T}_{\text{high}} = T_{\text{high}} \odot M_3(T_{\text{high}}).
\end{equation}

\textbf{Adaptive Frequency Gate.}
A learnable channel-wise gate $\alpha$ balances the two components based on their global statistics:
\begin{equation}
    \alpha = \sigma\!\left(\mathbf{W}_g \ast \text{GAP}(|[\tilde{T}_{\text{low}} \| \tilde{T}_{\text{high}}]|)\right),
\end{equation}
\begin{equation}
    T_{\text{final}} = \alpha \odot \tilde{T}_{\text{low}} + (1{-}\alpha) \odot \tilde{T}_{\text{high}}.
\end{equation}

\textbf{Safe Residual Fusion.}
A scalar confidence gate $s = \sigma(\mathbf{W}_2 \ast \text{ReLU}(\mathbf{W}_1 \ast \text{GAP}(R_i))) \in [0,1]$ estimates RGB reliability. The gate modulates only the RGB features entering a refinement branch—not the residual path—ensuring gradient flow is never disrupted:
\begin{equation}
    F_i = R_i + \phi_{\text{refine}}([s \cdot R_i \| T_{\text{final}}]).
    \label{eq:safe_residual}
\end{equation}
The correction $\Delta = \phi_{\text{refine}}(\cdot)$ can take negative values, canceling noisy RGB content when the thermal signal indicates this is necessary.

\begin{figure*}[!t]
    \centering
    \includegraphics[width=0.90\linewidth]{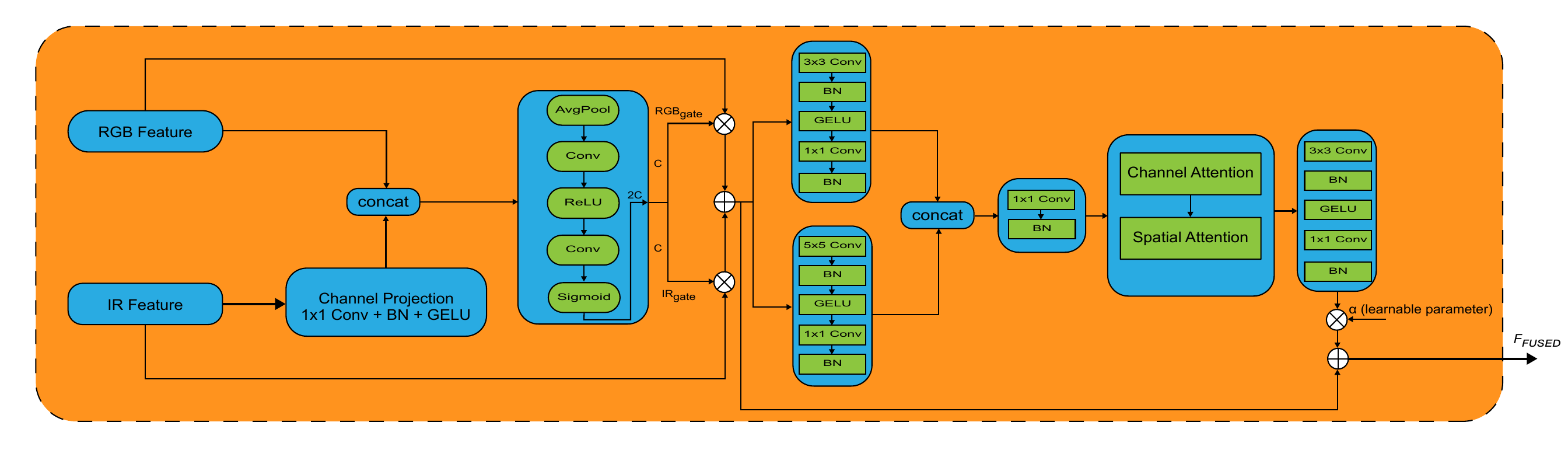}
    \caption{Semantic Fusion Module (stages 3--4).}
    \label{fig:semantic_fusion}
\end{figure*}

\subsection{Semantic Fusion Module}
\label{sec:semantic_fusion}

At deeper stages, features carry abstract semantic information where frequency decomposition is less meaningful (\cref{fig:semantic_fusion}). The thermal features are projected as before.

\textbf{Cross-Modal Channel Gating.}
RGB and projected thermal features are concatenated and compressed through two FC layers to produce a $2C$-dimensional gate, which is split into modality-specific gates $\mathbf{g}_{\text{rgb}}, \mathbf{g}_{\text{ir}}$:
\begin{equation}
    F_{\text{gated}} = \mathbf{g}_{\text{rgb}} \odot R_i + \mathbf{g}_{\text{ir}} \odot T_i'.
\end{equation}
The gates are not constrained to sum to one, allowing the network to amplify both modalities when both are informative.

\textbf{Multi-Scale Feature Extraction and Attention.}
The gated features are processed through parallel depthwise separable convolutions ($3{\times}3$ and $5{\times}5$) to capture objects at varying scales, then concatenated and fused via $1{\times}1$ convolution. Sequential SE-style~\cite{senet} channel attention and CBAM-style spatial attention refine the result. A learnable parameter $\gamma$ (initialized to $0.1$) scales the refinement branch before a residual connection:
\begin{equation}
    F_i = F_{\text{gated}} + \gamma \cdot \phi_{\text{refine}}(\text{SA}(\text{CA}(F_{\text{ms}}))).
\end{equation}

\subsection{PANet-Style Bidirectional Decoder}
\label{sec:decoder}

The decoder receives $\{F_i\}_{i=1}^{4}$ and produces segmentation logits through five stages. All features are first projected to a common dimension $D = C_1$ via $1{\times}1$ convolutions.

\textbf{Top-Down Path.}
Starting from the deepest level, each feature is upsampled and added to the next shallower lateral projection, then refined by a $3{\times}3$ convolution:
\begin{equation}
    P_i^{\text{td}} = \text{Conv}_3\!\left(L_i + \text{Up}(P_{i+1}^{\text{td}})\right), \quad P_4^{\text{td}} = L_4.
\end{equation}

\textbf{Bottom-Up Path.}
The direction is reversed: each level is downsampled via stride-2 convolution, concatenated with the corresponding top-down feature, and merged:
\begin{equation}
    P_i^{\text{bu}} = \text{Conv}_3\!\left([\text{Down}(P_{i-1}^{\text{bu}}) \| P_i^{\text{td}}]\right), \quad P_1^{\text{bu}} = P_1^{\text{td}}.
\end{equation}

\textbf{Global Context and Aggregation.}
A context vector $\mathbf{c} \in \mathbb{R}^{D}$ from $F_4$ modulates all bottom-up features via $(0.5 + 0.5 \cdot \sigma(\mathbf{c}))$, bounding the modulation between $0.5$ and $1.0$. All features are upsampled to the highest resolution, concatenated, and processed through a two-layer aggregation block and segmentation head.

\textbf{Deep Supervision.}
During training, a $1{\times}1$ classification head at each level provides auxiliary gradients with weights $\omega_i \in \{0.1, 0.2, 0.3, 0.4\}$ increasing with depth, reflecting that deeper levels with richer semantics benefit more from direct supervision.

\subsection{Loss Function}
\label{sec:loss}

The model is trained with a composite loss combining complementary objectives:
\begin{equation}
\mathcal{L} = \mathcal{L}_{\text{main}} + \lambda_{\text{aux}}\!\left(\mathcal{L}_{\text{aux}} + \sum_{i=1}^{4} \omega_i \mathcal{L}_{\text{deep}}^{(i)}\right)
\end{equation}

where $\mathcal{L}_{\text{main}}$ is a weighted sum of label-smoothed cross-entropy with inverse-frequency class weights, soft Dice loss~\cite{milletari2016vnet} for region-level overlap, Lov\'{a}sz-softmax~\cite{berman2018lovasz} for direct IoU optimization, OHEM cross-entropy~\cite{shrivastava2016ohem} for hard pixel mining, Focal loss~\cite{lin2017focal} for down-weighting easy examples, and a boundary loss using Laplacian-detected edges. $\mathcal{L}_{\text{aux}}$ is an auxiliary cross-entropy at a mid-level decoder feature, and $\mathcal{L}_{\text{deep}}$ sums per-level cross-entropy losses with the depth-increasing weights described above. All loss weights and training hyperparameters are detailed in \cref{sec:implementation}.

\section{Experimental Results and Analyses}
\label{sec:experiments}

\subsection{Datasets}

\textbf{MFNet}~\cite{mfnet} contains 1,569 RGB-thermal image pairs (820 daytime, 749 nighttime) of urban driving scenes at $480 \times 640$ resolution with 9 semantic classes, split into 784/392 training/testing pairs.

\textbf{PST900}~\cite{PST900} provides 894 aligned RGB-thermal pairs at $720 \times 1280$ resolution from diverse scenes with 5 classes, using the official 597/291 train/test split.

\subsection{Implementation Details}
\label{sec:implementation}

Both encoders use ConvNeXt V2~\cite{convnextv2} with FCMAE pre-training on ImageNet-22k. Training uses AdamW~\cite{adamw} with layer-wise learning rate decay ($\times 0.9$) and differential rates for backbone ($5{\times}10^{-5}$), fusion ($2{\times}10^{-4}$), and decoder ($3{\times}10^{-4}$). We train for 200 epochs with batch size 4, cosine annealing with warm restarts, mixed-precision, and EMA (decay 0.999). The composite loss weights are: CE ($0.4$), Dice ($0.2$), Lov\'{a}sz ($0.2$), OHEM ($0.1$), Boundary ($0.1$), Focal ($0.25$, $\gamma{=}2.5$). Inverse-frequency class weights with dataset-specific boosting are applied. At inference, we apply test-time augmentation (TTA) with horizontal flip. All experiments use a single NVIDIA RTX A6000 GPU.

\subsection{Comparison with State-of-the-Art}

\subsubsection{Results on MFNet}

\begin{figure}[t]
    \centering
    \includegraphics[width=\columnwidth]{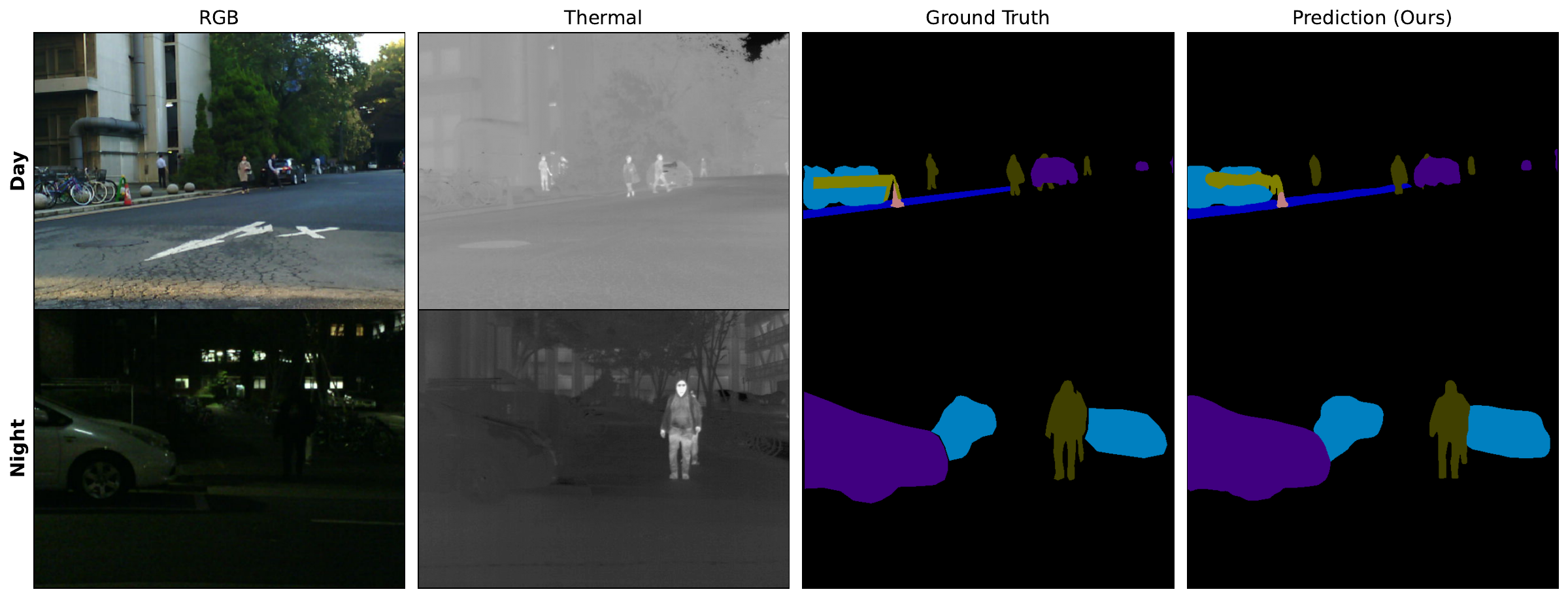}
    \caption{Qualitative results on MFNet day (top) and night (bottom) scenes. The model produces accurate segmentation in both conditions by leveraging thermal information.}
    \label{fig:daynight_qual}
\end{figure}

\Cref{tab:mfnet_sota} presents per-class and overall mIoU on the MFNet dataset. Our Nano variant achieves the highest mIoU (61.73\%) among all compared methods, using only \textbf{35.43M} parameters and \textbf{51.39 GFLOPs}---roughly $1.4\times$ fewer parameters and $1.7\times$ fewer FLOPs than Sigma-T~\cite{Sigma} (48.3M, 89.5G), and $3.4\times$ fewer parameters than CMNeXt~\cite{CMNext} (119.6M). Furthermore, our Tiny model (61.13\%) outperforms CMX~\cite{CMX} (59.7\%, 139.9M) and CMNeXt~\cite{CMNext} (59.9\%, 119.6M) with substantially fewer parameters. The Base model (59.63\%) shows reduced performance on MFNet due to a capacity--data mismatch: with 189M parameters trained on only 784 images, it exhibited training instability (NaN loss at epoch 71 of 200), reaching its best validation mIoU (59.7\%) at epoch 47 before diverging. In contrast, Tiny and Nano trained stably for all 200 epochs. This pattern reverses on PST900, where higher-resolution images ($720{\times}1280$) provide more pixel-level supervision and the Base model achieves the best result (88.48\%), confirming that performance scales with the data-to-parameter ratio.

Among frequency-aware methods, our approach surpasses both SGFNet~\cite{zhang2025spectral} (60.1\%) and Wavelet-CNet~\cite{waveletcnet} (58.3\%) while using a lighter backbone, confirming the effectiveness of IR-specific spatial-domain frequency decomposition.

\begin{table*}[t]
\centering
\caption{Quantitative comparisons on the MFNet day-night evaluation set (9 classes). Results with TTA (horizontal flip); see \Cref{tab:tta} for TTA-free baselines. Best in \textbf{bold}, second best \underline{underlined}.}
\label{tab:mfnet_sota}
\resizebox{\textwidth}{!}{%
\begin{tabular}{l|c|r|r|c|c|c|c|c|c|c|c|c|c}
\hline
Method & Backbone & Params (M) & FLOPs (G) & Unlabeled & Car & Person & Bike & Curve & Stop & Guardrail & Cone & Bump & mIoU \\
\hline
MFNet~\cite{mfnet} & --- & --- & --- & 96.9 & 65.9 & 58.9 & 42.9 & 29.9 & 9.9 & 0.0 & 25.2 & 27.7 & 39.7 \\
RTFNet~\cite{RTFNet} & ResNet-152 & 245.7 & 185.2 & 98.5 & 87.4 & 70.3 & 62.7 & 45.3 & 29.8 & 0.0 & 29.1 & 55.7 & 53.2 \\
ABMDRNet~\cite{ABMDRNet} & ResNet-50 & 64.6 & 194.3 & \textbf{98.6} & 84.8 & 69.6 & 60.3 & 45.1 & 33.1 & 5.1 & 47.4 & 50.0 & 54.8 \\
GMNet~\cite{GMNet} & ResNet-50 & 149.8 & 153.0 & 97.5 & 86.5 & 73.1 & 61.7 & 44.0 & 42.3 & \textbf{14.5} & 48.7 & 47.4 & 57.3 \\
EAEFNet~\cite{EAEFNet} & ResNet-152 & 200.4 & 147.3 & --- & 87.6 & 72.6 & 63.8 & 48.6 & 35.0 & \underline{14.2} & 52.4 & 58.3 & 58.9 \\
CMX~\cite{CMX} & MiT-B4 & 139.9 & 134.3 & 98.3 & 90.1 & 75.2 & 64.5 & 50.2 & 35.3 & 8.5 & 54.2 & 60.6 & 59.7 \\
CMNeXt~\cite{CMNext} & MiT-B4 & 119.6 & 131.9 & 98.4 & \textbf{91.5} & 75.3 & 67.6 & 50.5 & 40.1 & 9.3 & 53.4 & 52.8 & 59.9 \\
SGFNet~\cite{zhang2025spectral} & ResNet-152 & 163.9 & 249.2 & --- & 88.0 & 74.5 & 63.3 & 50.6 & 37.7 & 10.8 & 53.7 & \textbf{62.5} & 60.1 \\
Wavelet-CNet~\cite{waveletcnet} & ResNet-50 & 125.6 & 147.8 & --- & 87.8 & 71.9 & 62.8 & 48.4 & 41.8 & 10.1 & 53.7 & 50.7 & 58.3 \\
Sigma-T~\cite{Sigma} & VMamba-T & 48.3 & 89.5 & 98.4 & 90.8 & 75.2 & 66.6 & 48.2 & 38.0 & 8.7 & 55.9 & 60.4 & 60.2 \\
Sigma-S~\cite{Sigma} & VMamba-S & 69.8 & 138.9 & \underline{98.5} & \textbf{91.5} & \underline{75.8} & \underline{67.8} & 49.6 & 41.8 & 9.6 & 54.8 & 60.4 & 61.1 \\
Sigma-B~\cite{Sigma} & VMamba-B & 121.4 & 240.7 & \underline{98.5} & 91.1 & 75.2 & \textbf{68.0} & 50.8 & 43.0 & 9.7 & \underline{57.6} & 57.9 & \underline{61.3} \\
\hline
Ours (Base) & ConvNeXtV2-B & 189.1 & 242.8 & 98.3 & 90.9 & 74.9 & 65.7 & 47.7 & \underline{48.2} & 0.0 & \textbf{58.4} & 52.3 & 59.6 \\
Ours (Tiny) & ConvNeXtV2-T & 63.5 & 85.4 & 98.4 & \underline{91.1} & 75.5 & 66.4 & \textbf{53.2} & 46.2 & 0.0 & 57.9 & 61.1 & 61.1 \\
\textbf{Ours (Nano)} & ConvNeXtV2-N & \textbf{35.4} & \textbf{51.3} & 98.4 & 90.2 & \textbf{75.9} & 66.5 & \underline{51.6} & \textbf{53.9} & 0.0 & 57.0 & \underline{61.9} & \textbf{61.7} \\
\hline
\end{tabular}%
}
\end{table*}

\begin{figure*}[t]
    \centering
    \includegraphics[width=\textwidth]{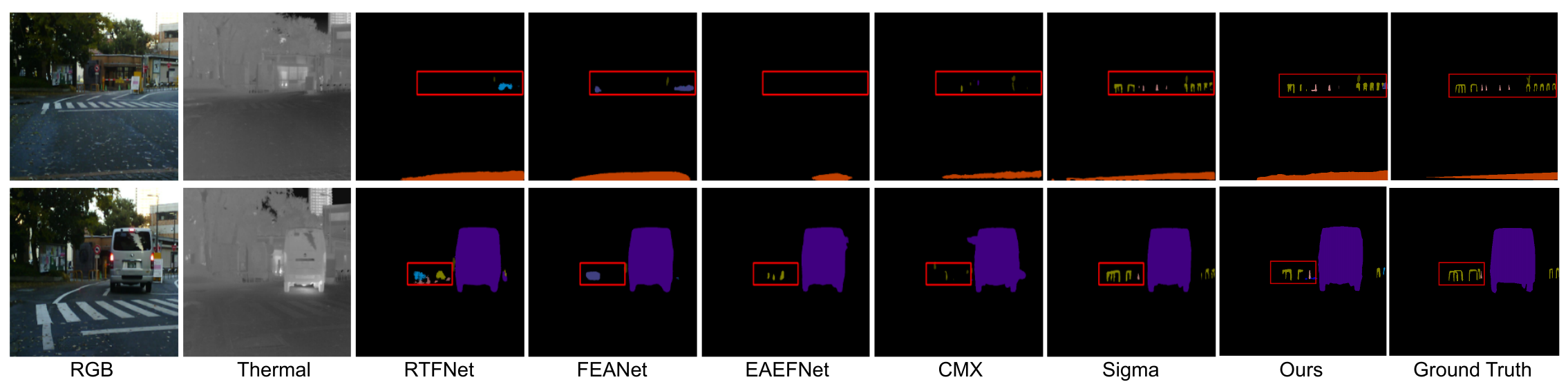}
    \caption{Qualitative comparison with existing methods on the MFNet dataset. Red boxes highlight regions of interest. Our method produces more complete and accurate segmentation, particularly for small and distant objects.}
    \label{fig:qualitative}
\end{figure*}
\subsubsection{Results on PST900}

\Cref{tab:pst900_sota} presents results on PST900. Unlike MFNet, performance scales with model capacity: our Base model reaches 88.48\% mIoU, competitive with Sigma-T~\cite{Sigma} (88.6\%) while both methods employ approximately similar computational budgets at this resolution. Our Tiny variant (87.88\%) outperforms CACFNet~\cite{CACFNet} (86.6\%, ConvNeXt-B) with a significantly smaller backbone.

\begin{table}[t]
\centering
\caption{Quantitative comparisons on PST900 (5 classes). Results with TTA (horizontal flip); see \Cref{tab:tta} for TTA-free baselines.}
\label{tab:pst900_sota}
\resizebox{\columnwidth}{!}{%
\begin{tabular}{l|c|c|c|c|c|c|c}
\hline
Method & Backbone & BG & Fire Ext. & Backpack & Hand-Drill & Survivor & mIoU \\
\hline
ABMDRNet~\cite{ABMDRNet} & ResNet-50 & 99.0 & 66.2 & 67.9 & 61.5 & 62.0 & 71.3 \\
GMNet~\cite{GMNet} & ResNet-50 & 99.4 & 73.8 & 83.8 & 85.2 & 78.4 & 84.1 \\
EGFNet~\cite{EGFNet} & ResNet-101 & 99.6 & 80.0 & \textbf{90.6} & 76.1 & 80.9 & 85.4 \\
CACFNet~\cite{CACFNet} & ConvNeXt-B & \textbf{99.6} & 82.1 & 89.5 & 80.9 & 80.8 & 86.6 \\
SGFNet~\cite{zhang2025spectral} & ResNet-152 & 99.5 & \underline{83.2} & 86.9 & 79.8 & 77.24 & 85.37 \\
Wavelet-CNet~\cite{waveletcnet} & ResNet-50 & 99.54 & 83.08 & 88.03 & 79.76 & 78.43 & 85.77 \\
Sigma-T~\cite{Sigma} & VMamba-T & \textbf{99.6} & 81.9 & 89.8 & \underline{88.7} & \underline{82.7} & \textbf{88.6} \\
Sigma-S~\cite{Sigma} & VMamba-S & \textbf{99.6} & 79.4 & 88.7 & \textbf{90.2} & 81.2 & 87.8 \\
\hline
\textbf{Ours (Base)} & ConvNeXtV2-B & \textbf{99.6} & \textbf{83.9} & 89.9 & 87.3 & 81.5 & \underline{88.4} \\
Ours (Tiny) & ConvNeXtV2-T & \textbf{99.6} & 79.4 & \underline{90.1} & 87.3 & \textbf{82.9} & 87.8 \\
Ours (Nano) & ConvNeXtV2-N & 99.5 & 78.3 & 88.3 & 84.6 & 80.2 & 86.2 \\
\hline
\end{tabular}%
}
\end{table}

\subsection{Ablation Studies}

To validate individual design choices, we conduct ablation experiments on MFNet using the Tiny backbone (\cref{tab:ablation}). We evaluate (i)~replacing the Semantic Fusion at stages 3--4 with the Frequency-Based Fusion Module (``All Freq.''), (ii)~disabling deep supervision, and (iii)~replacing the PANet decoder with a standard FPN decoder.

\begin{table}[t]
\centering
\caption{Ablation study on MFNet (ConvNeXtV2-Tiny). ``All Freq.'' uses Frequency-Based Fusion at all four stages.}
\label{tab:ablation}
\resizebox{\columnwidth}{!}{%
\begin{tabular}{l|c|c|c|cc|c}
\hline
Configuration & Decoder & DeepSup & Fusion (3--4) & w/o TTA & w/ TTA & Params (M) \\
\hline
\textbf{Full model} & PANet & \checkmark & Semantic & 60.88 & {61.13} & 63.57 \\
All Freq. fusion & PANet & \checkmark & Frequency & 61.16 & 61.18 & 75.40 \\
w/o Deep Supervision & PANet & --- & Semantic & 60.11 & 60.33 & 63.57 \\
FPN decoder & FPN & --- & Semantic & 60.85 & 61.10 & 64.31 \\
\hline
\end{tabular}%
}
\end{table}

\begin{figure}[t]
    \centering
    \includegraphics[width=\columnwidth]{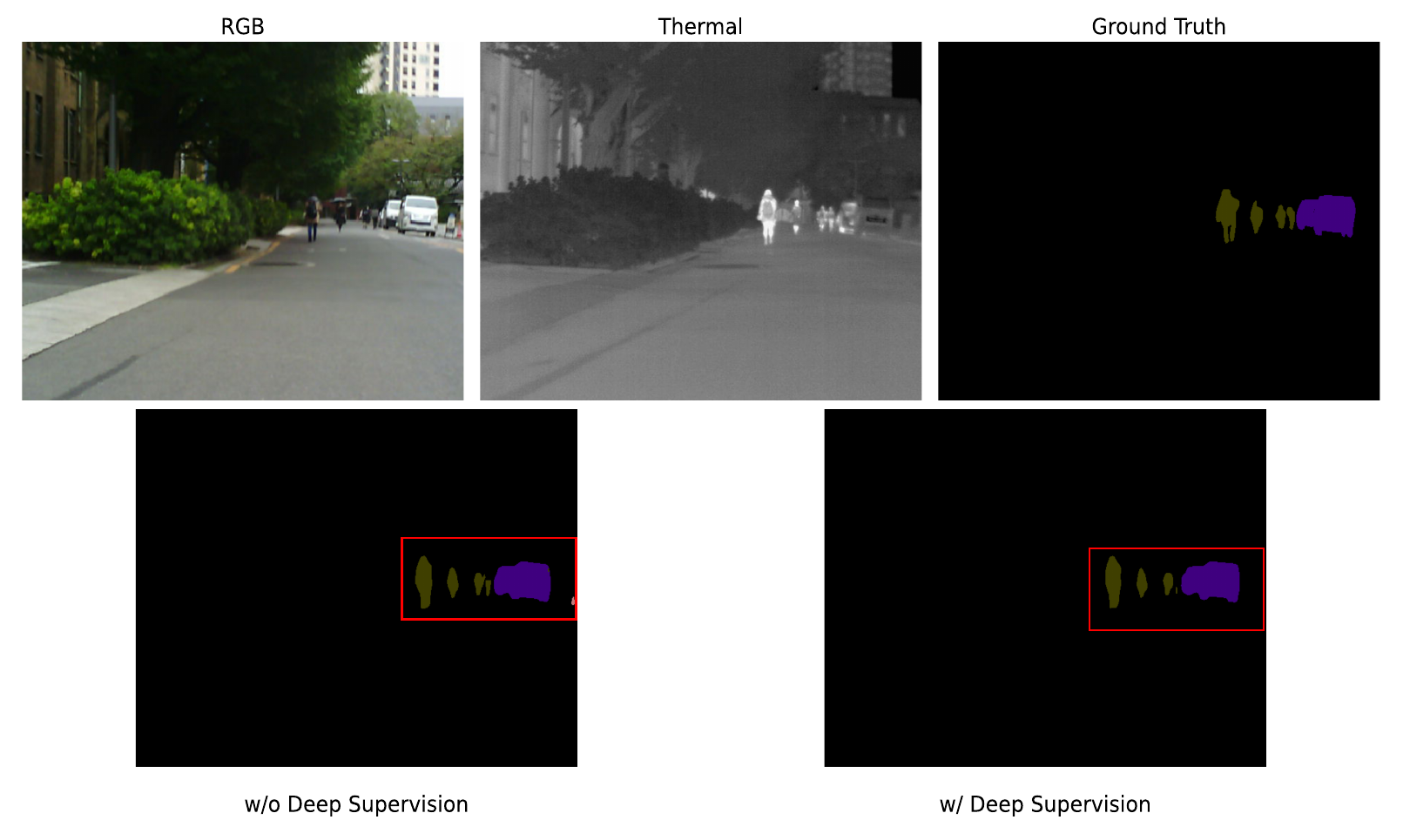}
    \caption{Effect of deep supervision. The model trained with deep supervision (right) produces more complete segmentation of distant and small objects compared to the model without (left).}
    \label{fig:deepsup}
\end{figure}

\textbf{Effect of Deep Supervision.}
Removing deep supervision causes the largest drop in mIoU ($-0.80\%$), validating its role in improving gradient flow through bidirectional aggregation paths. The improvement is most pronounced for underrepresented classes such as car stop and bump. \Cref{fig:deepsup} illustrates the qualitative difference: the model trained with deep supervision produces more complete segmentation of distant objects.

\textbf{Stage-Adaptive vs.\ Uniform Frequency Fusion.}
Using the Frequency-Based Fusion Module at all four stages (``All Freq.'') yields marginally higher mIoU (61.18\%) but at the cost of 18.6\% more parameters (75.40M vs.\ 63.57M) and 5.8\% more GFLOPs (90.33 vs.\ 85.42). Furthermore, the $0.05\%$ difference (61.18 vs.\ 61.13) falls within the typical run-to-run variation, while the parameter reduction ($-15.7\%$) is substantial and consistent. The stage-adaptive design achieves comparable accuracy more efficiently: the lighter Semantic Fusion is better suited for the reduced spatial resolution at deeper stages.

\textbf{PANet vs.\ FPN Decoder.}
Replacing PANet with FPN (including disabling deep supervision) yields 61.10\% mIoU, indicating that the bidirectional path provides a modest improvement primarily through its compatibility with deep supervision.

\textbf{Effect of TTA.}
\Cref{tab:tta} isolates the contribution of test-time augmentation.
TTA provides at most $+0.45\%$ mIoU and is \emph{negative} for two
model--dataset pairs (Base on MFNet, Nano on PST900), confirming that
performance gains are driven by the architecture rather than
test-time engineering.

\subsection{Day vs.\ Night Performance}

To evaluate robustness under varying illumination, we test on the MFNet day (437 images) and night (355 images) subsets separately (\cref{tab:daynight}).

\begin{table}[t]
\centering
\caption{Day/Night mIoU (\%) on MFNet test set.}
\label{tab:daynight}
\begin{tabular}{l|c|c}
\hline
Model & Day & Night \\
\hline
Ours (Tiny) & 55.46 & 60.84 \\
Ours (Nano) & 54.54 & 63.04 \\
\hline
\end{tabular}
\end{table}

\begin{table}[t]
\centering
\caption{Effect of TTA (horizontal flip) on mIoU (\%).}
\label{tab:tta}
\resizebox{\columnwidth}{!}{%
\begin{tabular}{l|cc|cc}
\hline
 & \multicolumn{2}{c|}{MFNet} & \multicolumn{2}{c}{PST900} \\
Model & w/o TTA & w/ TTA & w/o TTA & w/ TTA \\
\hline
Nano & 61.28 & 61.73 & 86.29 & 86.24 \\
Tiny & 60.88 & 61.13 & 87.53 & 87.88 \\
Base & 59.72 & 59.63 & 88.35 & 88.48 \\
\hline
\end{tabular}%
}
\end{table} 

Both models achieve higher mIoU at night than during the day, a pattern consistent with prior RGB-T methods~\cite{Sigma}. This confirms that the proposed fusion architecture effectively exploits thermal information when RGB is degraded. The Nano variant shows a particularly large night advantage ($+8.50\%$), suggesting that its compact encoder focuses more strongly on thermal cues in the absence of reliable visible-light features. \Cref{fig:daynight_qual} shows representative day and night predictions, where the thermal modality enables accurate segmentation even under severe illumination degradation.

\subsection{Discussion}

Our method achieves competitive mIoU across both benchmarks while maintaining a favorable efficiency profile. On MFNet, the Nano variant (35.43M, 51.39G) achieves 61.73\% mIoU with $1.4\times$ fewer parameters and $1.7\times$ fewer FLOPs than Sigma-T (48.3M, 89.5G, 60.2\%), while using $3.4\times$ fewer parameters than CMNeXt (119.6M, 59.9\%). On PST900, our Base model (88.48\%) approaches HAPNet~\cite{hapnet} (89.0\%) while operating within a purely convolutional framework without requiring transformer-specific positional encodings. \Cref{fig:qualitative} provides qualitative comparisons with  RTFNet~\cite{RTFNet}, FEANet~\cite{FEANet}, EAEFNet~\cite{EAEFNet}, CMX~\cite{CMX}, and Sigma~\cite{Sigma} showing that our method achieves more complete segmentation particularly for small objects and boundary regions.

A notable limitation is the persistent failure on the Guardrail class (0.0\% IoU across all variants), a difficulty shared by most methods in the literature (\eg, RTFNet 0.0\%, CMX 8.5\%, Sigma-B 9.7\%). Beyond the extreme class rarity---only 66 of 1,568 training images contain guardrail pixels---the root cause is a severe train/test domain shift: 97\% of training guardrails appear in daytime images, while \emph{all} test guardrails appear exclusively at night. This distributional mismatch makes the class nearly impossible to learn under standard training protocols. Future work may address this via domain-aware augmentation or night-specific synthetic data generation.
\section{Conclusion}
\label{sec:conclusion}

We presented a hierarchical fusion architecture for RGB-thermal semantic segmentation that employs stage-adaptive fusion strategies tailored to the nature of features at different encoding depths. At early stages, a Frequency-Based Fusion Module decomposes infrared features into low- and high-frequency components and selectively enhances RGB representations through a confidence-gated residual mechanism. At deeper stages, a Semantic Fusion Module establishes cross-modal correspondences through channel gating and multi-scale attention. A PANet-style bidirectional decoder with depth-weighted deep supervision aggregates the fused multi-scale features into the final segmentation.

Experiments on MFNet and PST900 demonstrate that our approach achieves competitive results---61.73\% and 88.48\% mIoU, respectively---while maintaining a favorable efficiency profile. Our smallest variant (35.43M parameters, 51.39 GFLOPs) outperforms substantially larger models, and ablation studies confirm the contributions of deep supervision and the stage-adaptive fusion design. The strong nighttime performance further validates the effectiveness of the proposed IR-centric frequency decomposition for exploiting thermal information when RGB is degraded.

Future work includes addressing the extreme class imbalance and domain shift observed for rare classes such as guardrail, exploring lightweight transformer-based encoders, and extending the frequency-guided fusion paradigm to other multi-modal perception tasks.

{
    \small
    \bibliographystyle{ieeenat_fullname}
    \bibliography{main}
}


\end{document}